\documentclass[letterpaper]{article} 
\usepackage{aaai2026}  
\usepackage{times}  
\usepackage{helvet}  
\usepackage{courier}  
\usepackage[hyphens]{url}  
\usepackage{graphicx} 
\usepackage{natbib}  
\usepackage{caption} 
\usepackage{algorithm}
\usepackage{algorithmic}

\usepackage{amsmath}
\usepackage{amssymb}
\usepackage{mathtools}
\usepackage{amsthm}

\usepackage[utf8]{inputenc} 
\usepackage[T1]{fontenc}    
\usepackage{url}            
\usepackage{booktabs}       
\usepackage{amsfonts}       
\usepackage{nicefrac}       
\usepackage{microtype}      
\usepackage{xcolor}         

\newcommand{\method}{{FUEL}\xspace}

\definecolor{sunwoogreen}{RGB}{32, 200, 150}
\definecolor{sunwoogreen2}{RGB}{67, 148, 58}
\definecolor{sunwooyellow}{rgb}{1.0, 1.0, 0.0}
\definecolor{sunwooyellow2}{RGB}{228, 208, 10}

\newcommand{\std}{\scriptsize}
\newcommand{\best}{\cellcolor{sunwoogreen}}  
\newcommand{\secb}{\cellcolor{sunwooyellow}}  

\usepackage{amsmath}
\DeclareMathOperator*{\argmax}{arg\,max}

\usepackage{comment}
\usepackage{xspace}
\usepackage{xcolor}
\usepackage{booktabs}

\usepackage{graphicx}

\usepackage{graphics}
\usepackage{xspace}

\usepackage{booktabs}
\usepackage{soul}

\usepackage{float}
\usepackage{amsmath}
\usepackage{amssymb}
\usepackage{mathtools}
\usepackage{amsfonts}
\usepackage{amsbsy}
\usepackage{amsthm}

\usepackage{multirow}
\usepackage[mathscr]{eucal} 
\usepackage{mathtools}
\usepackage{verbatim}
\usepackage{tabularx}
\usepackage{enumitem}
\usepackage{array}
\usepackage{soul}
\usepackage{dsfont}
\usepackage{url}

\usepackage[most]{tcolorbox}

\usepackage{pifont}
\usepackage{colortbl}
\usepackage[svgnames,x11names,table]{xcolor}
\usepackage[inkscapeformat=png]{svg}
\usepackage{makecell}
\usepackage{xspace}
\usepackage{footmisc} 

\definecolor{googleblue}{RGB}{66,133,244}
\definecolor{googlered}{RGB}{219,68,55}
\definecolor{googleyellow}{RGB}{244,180,0}
\definecolor{googlegreen}{RGB}{15,157,88}

\usepackage[capitalize,noabbrev]{cleveref}

\theoremstyle{plain}
\newtheorem{theorem}{Theorem}

\theoremstyle{definition}

\theoremstyle{remark}

\usepackage{newfloat}
\usepackage{listings}
\DeclareCaptionStyle{ruled}{labelfont=normalfont,labelsep=colon,strut=off} 
\floatstyle{ruled}
\newfloat{listing}{tb}{lst}{}
\floatname{listing}{Listing}
\makeatletter
\renewcommand{\@seccntformat}[1]{\csname the#1\endcsname\quad}
\makeatother

\title{Feature-Centric Unsupervised Node Representation Learning \\ Without Homophily Assumption}
\author{
    Sunwoo Kim\textsuperscript{\rm 1}, 
    Soo Yong Lee\textsuperscript{\rm 1}, 
    Kyungho Kim\textsuperscript{\rm 1}, 
    Hyunjin Hwang\textsuperscript{\rm 1}, 
    Jaemin Yoo\textsuperscript{\rm 2}, 
    Kijung Shin\textsuperscript{\rm 1,2}
}

\affiliations{
    \textsuperscript{\rm 1}Kim Jaechul Graduate School of AI, KAIST, Seoul, South Korea \\
    \textsuperscript{\rm 2}School of Electrical Engineering, KAIST, Daejeon, South Korea \\
    \{kswoo97, syleetolow, kkyungho, hyunjinhwang, jaemin, kijungs\}@kaist.ac.kr
}

\usepackage{bibentry}

\begin{document}

\maketitle

\begin{abstract}
Unsupervised node representation learning aims to obtain meaningful node embeddings without relying on node labels.
To achieve this, graph convolution, which aggregates information from neighboring nodes, is commonly employed to encode node features and graph topology.
{However, excessive reliance on graph convolution can be suboptimal—especially in non-homophilic graphs—since it may yield unduly similar embeddings for nodes that differ in their features or topological properties. 
As a result, adjusting the degree of graph convolution usage has been actively explored in supervised learning settings, whereas such approaches remain underexplored in unsupervised scenarios.}
To tackle this, we propose FUEL, which adaptively learns the adequate degree of graph convolution usage by aiming to enhance intra-class similarity and inter-class separability in the embedding space. 
Since classes are unknown, FUEL leverages node features to identify node clusters and treats these clusters as proxies for classes.
Through extensive experiments using 15 baseline methods and 14 benchmark datasets, we demonstrate the effectiveness of FUEL in downstream tasks, achieving state-of-the-art performance across graphs with diverse levels of homophily.
\end{abstract}

\section{Introduction}
\label{sec:intro}
Node embeddings, vector representations that capture the corresponding node information, enable the use of various machine learning models in tackling downstream tasks such as node classification and clustering.
Among various approaches, unsupervised node representation learning—designed to obtain useful embeddings without label supervision—has gained significant attention~\citep{grover2016node2vec, hou2022graphmae, chen2024polygcl, ju2022multi}.
This approach is cost-effective, as it eliminates the need for labeling efforts and often outperforms supervised methods in label-scarce scenarios~~\citep{velivckovic2018deep}.

To capture input node features and topology, \textit{graph convolution}~\citep{kipf2016semi,wu2019simplifying, lee2024feature}, which aggregates and propagates features from a node's local neighborhood, is widely leveraged~\citep{he2023contrastive, wang2024hetergcl}. 
Especially, it has demonstrated its effectiveness in homophilic graphs, where similar nodes are likely to be connected~\citep{guo2023architecture}.

However, \textit{excessively relying} on graph convolution can be suboptimal, particularly in non-homophilic graphs where dissimilar nodes are frequently connected.
This limitation arises since graph convolution inherently promotes similarity in the embeddings of adjacent nodes, leading to highly similar embeddings for dissimilar nodes that are frequently linked.

While adjusting the degree of graph convolution usage has been widely studied in supervised settings, it remains largely underexplored in unsupervised scenarios.
In supervised settings, specialized GNN architectures learn an adequate degree of graph convolution based on the downstream supervision.
However, such supervision is absent in unsupervised learning settings, making it challenging to adjust the adequate degree of graph convolution usage.

To address this,  we propose \textbf{\method} (\textbf{\underline{F}}eature-centric \textbf{\underline{U}}nsupervised node r\textbf{\underline{E}}presentation \textbf{\underline{L}}earning), a novel unsupervised node representation learning method designed without homophily assumption.
In a nutshell, \method adaptively determines the degree of graph convolution usage to increase intra-class similarity and inter-class separability in the embedding space.
Since explicit class information is unavailable, \method relies instead on node features, which offer class-relevant information independent of the graph topology. 
Specifically, nodes with similar features are expected to belong to the same class, naturally forming a distinct cluster in the feature space.
Building on this basis and a specialized clustering scheme, \method learns the degree of graph convolution that coheres each cluster while separating different clusters—a strategy backed by both theoretical and empirical evidence. 
Subsequently, \method refines node embeddings from the selected level of graph convolution to further enhance separability among clusters.

Through extensive experiments including 15 baseline methods and 14 real-world benchmark graph datasets, we demonstrate the effectiveness of \method in two widely studied node-level downstream tasks: node classification and clustering. 
Notably, the superiority of \method holds across graphs with various graph-level homophily, achieving the best node classification and clustering performance on both the least homophilic (Texas) and the most homophilic (Photo) graphs.
Our key contributions are summarized as follows:
\begin{itemize}[leftmargin=*]    
    \item We introduce a clustering-based proxy for class separability applicable in unsupervised settings and demonstrate its effectiveness both empirically and theoretically (Section~\ref{sec:analysis}).
    \item We present \method, a novel unsupervised node representation learning method that adaptively learns the adequate degree of graph convolution through clustering (Section~\ref{sec:method}).
    \item We validate the effectiveness of \method through extensive experiments, demonstrating its superiority over baseline methods in 10 out of 14 settings (Section~\ref{sec:experiment}).
\end{itemize}
For reproducibility, code and datasets are available at~\url{https://github.com/kswoo97/unsupervised-non-homophilic}.

\section{Related work and preliminary}
\label{sec:relatedwork}

In this section, we provide the preliminaries of our research and review the relevant literature.

\subsection{Preliminary}

A graph $\mathcal{G} = (\mathcal{V},\mathcal{E}, \mathbf{X})$ is defined by a set of nodes $\mathcal{V}=\{v_{1},v_{2},\cdots,v_{\vert \mathcal{V}\vert}\}$, a set of edges $\mathcal{E}\subseteq\binom{\mathcal{V}}{2}$, and a node feature matrix $\mathbf{X} \in \mathbb{R}^{\vert \mathcal{V}\vert \times d}$.
Each edge $e_{j}\in \mathcal{E}$ is a node pair, and edges can also be represented by an adjacency matrix $\mathbf{A}\in \{0,1\}^{\vert \mathcal{V}\vert \times \vert \mathcal{V}\vert}$, where $\mathbf{A}_{i,j}=1$ if and only if $\{v_{i},v_{j}\}\in \mathcal{E}$.
Each node $v_{i} \in \mathcal{V}$ is equipped with a feature vector $\mathbf{x}_{i} \in \mathbb{R}^{d}$, which together form the node feature matrix $\mathbf{X}$, where the $i-$th row of $\mathbf{X}$ is $\mathbf{x}_{i}$.
Therefore, a graph can alternatively be written as $\mathcal{G}=(\mathbf{X},\mathbf{A})$.

\subsection{Related work}

\noindent\textbf{Unsupervised node representation learning.}
Early unsupervised node representation learning methods primarily focus on encoding the topological structure of a given graph~\citep{perozzi2014deepwalk, grover2016node2vec}. 
On the other hand, recent approaches focus on the integration of node features and graph topology through self-supervised learning~\citep{velivckovic2018deep, thakoor2021bootstrapped}.
Typically, they use (1) contrastive approaches that contrast nodes across different views~\citep{hassani2020contrastive, you2020graph} or (2) generative approaches that reconstruct masked node features~\citep{hou2022graphmae,hou2023graphmae2} and/or edges~\citep{li2023s, kim2024hypeboy}.
{Recent work adapts these approaches to perform well even in non-homophilic graphs, where nodes with dissimilar features and/or structural properties are often connected.~\citep{liu2023beyond, wang2024hetergcl}.  
Some of them leverage multiple graph filters and contrast the embeddings obtained through different filters~\citep{chen2024polygcl}.
Another line of work focuses on filtering non-homophilic edges using feature and/or topological similarity, followed by the application of existing self-supervised learning techniques~\citep{yang2024graph}.}

\noindent\textbf{Adjusting the degree of graph convolution usage.}
{In (semi-)supervised graph learning, adjusting the degree of graph convolution has been widely explored, often through the design of specialized graph neural networks (GNNs)~\citep{zheng2022graph, yu2024lg}.
Notably, GPR-GNN~\citep{chien2020adaptive} and GADC~\citep{luan2022revisiting} achieve this by learning a distinct coefficient for each $k$-th power of the normalized adjacency matrix used in graph convolution.
However, since these coefficients are tuned by minimizing downstream task losses (e.g., cross-entropy for node classification), learning the adequate degree of graph convolution usage remains largely unexplored in unsupervised settings.}

\noindent\textbf{Clustering-based learning.} 
Clustering, a process of grouping data based on their characteristics, has effectively guided unsupervised, particularly self-supervised, representation learning in computer vision~\citep{khorasgani2022slic, walawalkar2025videoclusternet}. 
These approaches typically cluster visually similar images and enable image encoders to capture these similarities through pseudo-label classification~\citep{asano2019self} and/or contrastive learning~\citep{caron2020unsupervised}.

\begin{figure*}[!t] 
    \centering
    \includegraphics[width=0.75\linewidth]{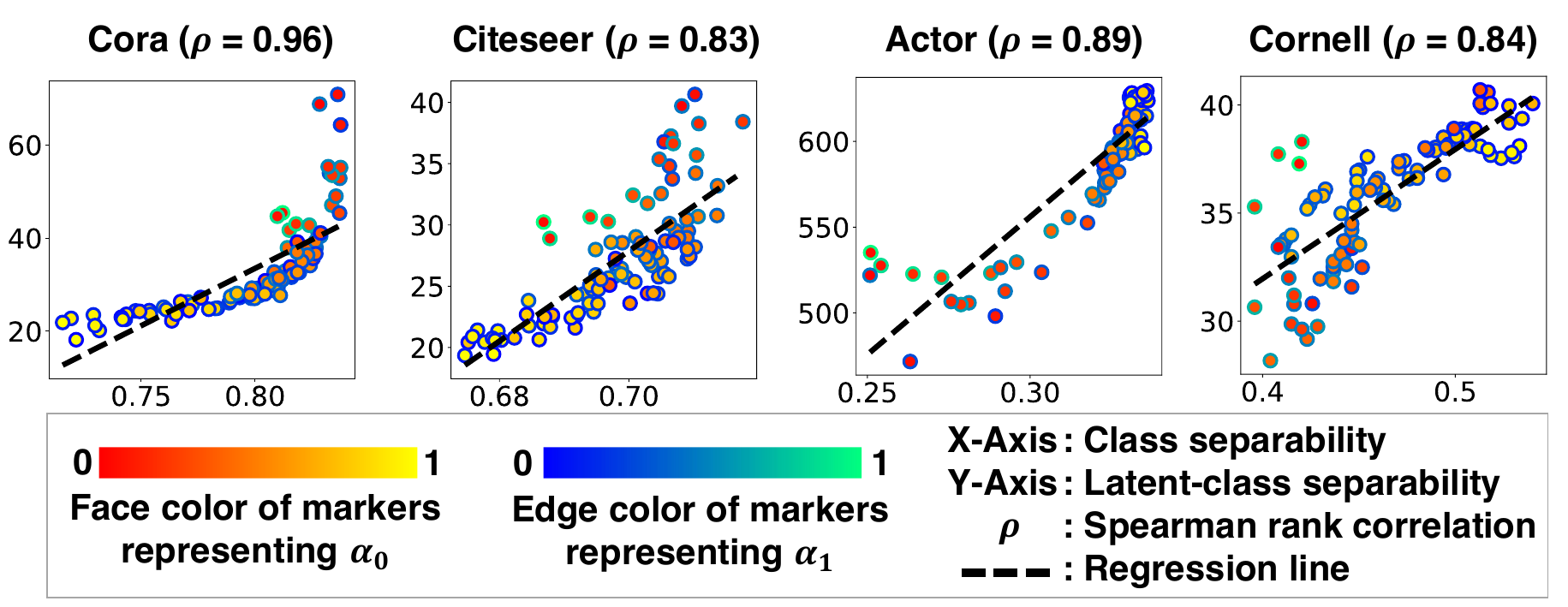}
    \caption{\label{fig:motivation}
    Effectiveness of latent-class separability as a proxy for class separability.
    Varying the graph convolution coefficients induces different degrees of graph convolution usage in the embeddings, which in turn leads to varying class separability. 
    Notably, latent-class separability, our proposed proxy, strongly correlates with the actual class separability.}
\end{figure*}

\section{Proposed proxy for class separability}
\label{sec:analysis}

Recall that the core idea of \method, our proposed method, is to learn an adequate degree of graph convolution usage that increases the embeddings' ability to distinguish classes, which we call \textit{class separability}.
In this section, we introduce its proxy: \textit{latent-class separability}.
We begin by presenting the high-level motivation and concept of latent-class separability (Section~\ref{subsec:motivation}), and then demonstrate its effectiveness as a proxy for class separability (Section~\ref{subsec:proxyjustification}).

\subsection{Motivation}\label{subsec:motivation}
\textbf{Goal and challenge.}
{As discussed above, we regard adequate graph convolution usage as one that improves class separability in the embedding space.
This property makes the learned embeddings well-suited for various popular downstream tasks, including node classification and clustering.
However, since we consider unsupervised settings, we are not given any node labels. 
Therefore, we need a proxy measure for class separability.}

\noindent\textbf{Proposed proxy.}
We tackle this challenge by focusing on node features, which inherently reflect node classes.
Nodes of the same class tend to share similar features, while those from different classes exhibit distinct ones.
Therefore, nodes naturally form clusters in the feature space, which we term \textit{latent classes}.
With an adequate degree of graph convolution, features within the same class likely become more similar, enhancing the separability of latent classes.
Based on this, we claim that \textit{latent-class separability}—the degree to which each latent class is internally cohesive and distinct from others—can serve as an effective proxy for class separability.

\subsection{Effectiveness of latent-class separability}\label{subsec:proxyjustification}
{In this subsection, we first empirically show, using real-world graph datasets, that latent-class separability can serve as an effective proxy for class separability. 
We then give a theoretical justification for using latent-class separability as a proxy for class separability by proving that, under some conditions, latent-class separability is equivalent to class separability.}

\noindent\textbf{Empirical analysis setup.}
We analyze the correlation between class separability and latent-class separability across different degrees of graph convolution usage.
{To this end, we control the degree through the following graph convolution function, which produces node embeddings $\mathbf{Z} \in \mathbb{R}^{\vert \mathcal{V}\vert \times d}$ as: $\mathbf{Z} = \alpha_{0}\mathbf{X} + \alpha_{1}\hat{\mathbf{A}}\mathbf{X} + \alpha_{2}\hat{\mathbf{A}}^{2}\mathbf{X}$, where $\hat{\mathbf{A}}$ and $\hat{\mathbf{A}}^{2}$ denote the row-wise normalized versions of $\mathbf{A}$ and $\mathbf{A}^{2}$, respectively; 
the coefficients satisfy $0 \leq \alpha_{i} \leq 1, \forall i \in \{0,1,2\}$, and $\sum_{i=0}^{2} \alpha_{i} = 1$.
We obtain 100 embeddings with varying degrees of graph convolution by, for each embedding: (1) sampling three scalars uniformly at random from $[0,1]$, (2) normalizing them by dividing each by their sum, and (3) assigning the resulting values to $\alpha_0$, $\alpha_1$, and $\alpha_2$, respectively.
In Section D.6 of the supplementary material, we further analyze representative graph convolution functions.}\footnote{We analyze APPNP~\cite{gasteiger2018predict} and GPR-GNN~\cite{chien2020adaptive}, and the results obtained using them are consistent with those presented in Section~\ref{subsec:proxyjustification}.
Detailed results are in Section D.6 of the supplementary material.}

We measure the class separability of an embedding by training a linear classifier on that embedding and reporting its accuracy on test nodes; {additional metrics that directly use node labels are in Appendix D.11 of~\cite{supplementary}, yielding the same result.}
In contrast, we measure the latent-class separability by using the Calinski–Harabasz index~\citep{calinski1974dendrite}, an unsupervised cluster quality metric that measures the ratio of mean inter-latent-class distance to mean intra-latent-class distance. 
Details on the Calinski–Harabasz index and its computation are provided in {Section D.2 of the supplementary material}.
We use two homophilic graphs (Cora and Citeseer) and two non-homophilic graphs (Actor and Cornell), whose details are provided in Section B of the supplementary material. 
Nodes are split into 10\% for training and 90\% for testing.

\noindent\textbf{Empirical analysis result.}
As shown in Figure~\ref{fig:motivation}, the latent-class separability of the embeddings is strongly and positively correlated with their class separability.
Specifically, across both homophilic and non-homophilic datasets, (1) Spearman rank correlation values exceed 0.82, and (2) the fitted linear regression lines have positive slopes.
These results suggest that latent-class separability is an effective proxy for class separability in real-world graph datasets.

\noindent\textbf{Theoretical analysis setup.}
We theoretically show that under certain conditions, if one degree of graph convolution usage yields greater class separability than another, the same ordering holds for their latent-class separability.
We consider two non-empty, disjoint node sets $\mathcal{C}_{0} \cup \mathcal{C}_{1} = \mathcal{V}$ of equal size (i.e., $\vert \mathcal{C}_{0} \vert = \vert \mathcal{C}_{1} \vert$), with each set corresponding to a distinct class.
Node features $x_{i}, \forall v_{i}\in \mathcal{V}$ are generated from a Gaussian distribution, based on the class to which each node belongs.
\footnote{{While our theoretical analysis focuses on two uniform-sized classes for an intuitive and rigorous theoretical analysis, we empirically demonstrate in Figure~\ref{fig:motivation} and Appendix D.10 of~\cite{supplementary} that latent-class separability remains an effective proxy for class separability in multi-class, imbalanced real-world graphs.}}
Formally, the following holds: $x_{i} \sim \mathcal{N}(\mu ,\sigma^{2}), \forall v_{i} \in \mathcal{C}_{0}$ and $x_{j} \sim \mathcal{N}(-\mu ,\sigma^{2})$ for $v_{j} \in \mathcal{C}_{1}$.
Each node $v_{i}$ has $n$ neighbors, denoted by $N(v_{i}) \subset\mathcal{V}$, of which $n_{0}$ belong the same class of $v_{i}$ and $n_{1}=n-n_{0}$ belong to a different class.
We define the embedding of $v_{i}$ obtained via graph convolution as $z_{i} = \left((1-w)x_{i} + w\frac{1}{n}\left(\sum_{v_{j} \in N(v_{i})} x_{j}\right)\right)$, where $w \in [0,1]$ denotes the degree of graph convolution usage.

We denote the class separability of $z$ as $\textit{\texttt{CS}}(z;n,n_0,w)$, which is defined as one minus the Bayes error rate of the Bayes classifier~\citep{bishop2006pattern} using the corresponding feature as an input 
(i.e., $\mathbb{E}_{x}[\mathbb{I}[P(v_{i}\in \mathscr{C}(v_{i}) \vert z_i )>P(v_{i}\notin \mathscr{C}(v_{i}) \vert z_i)]]$, where $\mathscr{C}(v_{i})$ is the class of $v_{i}$ and $P(v_{i} \in \mathcal{C} \vert x_{i})$ represents the probability assigned by the Bayes classifier that $v_{i}$ belongs to the class $\mathscr{C}(v_{i})$, given its embedding $z_{i}$).
In addition, we denote the latent-class separability of $z$ as $\texttt{LCS}(z;n,n_{0},w)$, which extends the Calinski-Harabasz index~\citep{calinski1974dendrite} as follows:
\begin{equation}
    \frac{(\mathbb{E}_{x}[z_{i} \vert v_{i} \in \mathcal{C}_{0}] - \mathbb{E}_{x}[z_{i} \vert v_{i} \in \mathcal{C}_{1}])^{2}}{\frac{\mathbb{E}_{x}[(z_{i} - \mathbb{E}[z_{i} \vert v_{i}\in \mathcal{C}_{0}])^{2} \vert v_{i} \in \mathcal{C}_{0}] + \mathbb{E}_{x}[(z_{i} - \mathbb{E}[z_{i} \vert v_{i}\in \mathcal{C}_{1}])^{2} \vert v_{i} \in \mathcal{C}_{1}]}{2}}.
\end{equation}

\noindent\textbf{Theoretical analysis result.}
Our theoretical finding is summarized in the following theorem:
\begin{theorem}[Effectiveness of latent-class separability]\label{thm:effectiveness}
If $n_{0}=n$ or $w,w' \in [\max(\frac{n-2n_{0}}{2n-2n_{0}}, 0), 1]$ hold, the following holds:
$\texttt{CS}(z;n,n_0,w) > \texttt{CS}(z;n,n_0,w')$ if and only if $\texttt{LCS}(z;n,n_0,w) > \texttt{LCS}(z;n,n_0,w')$.
\end{theorem}
\begin{proof}
    Refer to Section A of~\cite{supplementary}.
\end{proof}
As shown in Theorem~\ref{thm:effectiveness}, the ordering of class separability between two degrees of graph convolution usage holds if and only if the same ordering holds for their latent-class separability, given that certain conditions for $w$ and $w'$ are met.
This theoretical result supports the effectiveness of latent-class separability as a proxy for class separability, indicating that the degree of graph convolution usage leading to better class separability can be effectively inferred from latent-class separability—especially in unsupervised settings.

\section{Proposed method}
\label{sec:method}

\begin{figure*}[!t] 
    \centering
    {\includegraphics[width=0.8\linewidth]{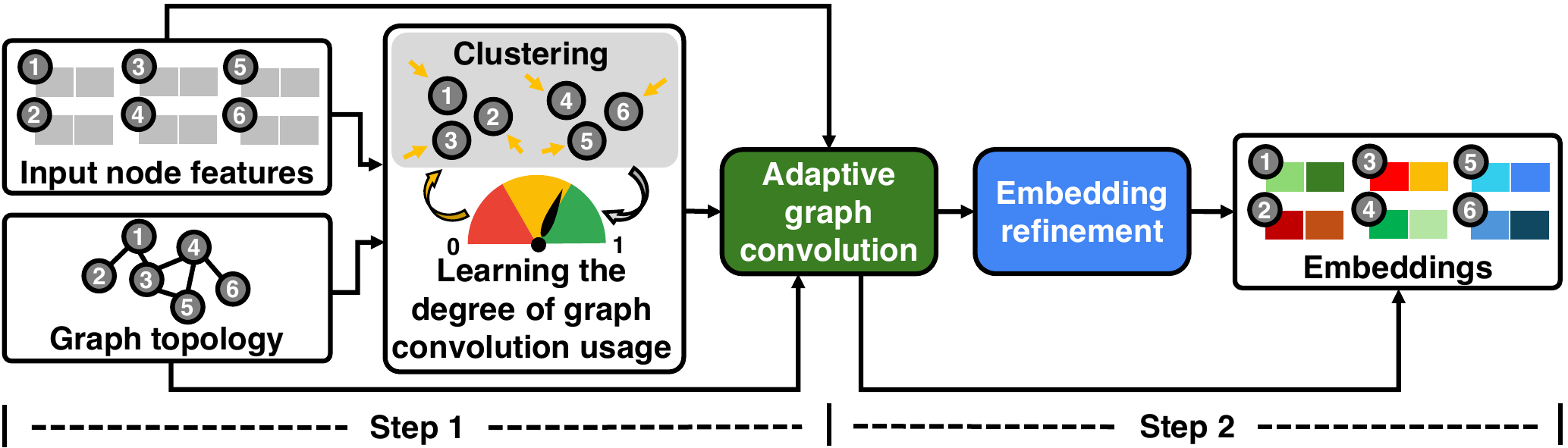}}
    \caption{
    {Overview of \method.} \method consists of two steps. 
    In Step 1, \method learns an adequate degree of graph convolution usage through the proposed clustering scheme and an adaptive graph convolution model.
    In Step 2, \method improves the cohesion of each latent class while preserving its distinction from others through embedding refinement.}
    \label{fig:method}
\end{figure*}

In this section, we introduce \method (\textbf{\underline{F}}eature-centric \textbf{\underline{U}}nsupervised node r\textbf{\underline{E}}presentation \textbf{\underline{L}}earning), a novel unsupervised node representation learning method. 
Building on the analyses provided in Section~\ref{sec:analysis}, \method learns the appropriate degree of graph convolution usage by aiming to enhance the latent-class separability.
To this end, \method leverages clusters identified by a specialized clustering scheme (Section~\ref{subsec:step1}) as latent classes.
Then, \method further enhances the latent-class separability of node embeddings from the selected degree of graph convolution usage through an additional refinement step (Section~\ref{subsec:step2}). 
Refer to Figure~\ref{fig:method} for a pictorial overview of \method.

\subsection{Step 1. A specialized clustering scheme}\label{subsec:step1}
\method employs a specialized clustering scheme to learn the adequate degree of graph convolution usage. 
In this scheme, an adaptive graph convolution model is employed, containing learnable parameters that directly control the impact of graph convolution on the resulting node embeddings.
These learnable parameters are optimized to maximize the separability of clusters in the resulting embedding space.

\noindent\textbf{Adaptive graph convolution model.}
As an adaptive graph convolution model, we use Eq.~\eqref{eq:adaptgc}, where a learnable weight is assigned to each graph-convolution-based embedding term: 
\begin{equation}\label{eq:adaptgc}
    \mathbf{X}^{*} = (\alpha_{0}\mathbf{X} + \alpha_{1}\tilde{\mathbf{A}}\mathbf{X} + \alpha_{2}\tilde{\mathbf{A}^{2}}\mathbf{X}) \in \mathbb{R}^{\vert \mathcal{V} \vert \times d},
\end{equation}
where $\alpha_{0},\alpha_{1},\alpha_{2} \in [0,1]$ and $\alpha_{0} + \alpha_{1} + \alpha_{2} = 1$ hold, and {$\mathbf{X}$ denotes the node feature matrix provided by the dataset}.
Note that Eq.~\eqref{eq:adaptgc} has the same functional form of the graph convolution function used in Section~\ref{sec:analysis}.
We learn three scalars $c_{0},c_{1},c_{2} \in \mathbb{R}$, which become $\alpha_{0},\alpha_{1},\alpha_{2} \in \mathbb{R}$ via the softmax function (i.e., $\alpha_{k} = \frac{\exp(c_{k})}{\sum^{2}_{j=0} \exp(c_{j})}, \forall k \in \{0, 1, 2\}$).

\noindent\textbf{Training for clustering.}
Our goal is to optimize the learnable weights in Eq.~\eqref{eq:adaptgc}.
To this end, inspired by entropy-based deep clustering methods in computer vision~\citep{ghasedi2017deep, zhao2023incremental, huang2020deep}, we incorporate entropy-based losses designed to achieve the following objectives: 
ensuring each node has a high cluster-assignment score to a single cluster ($\mathcal{L}_{1}$), 
and {encouraging the expected cluster assignment distribution to be close to uniform over clusters} to prevent nodes from collapsing into a single cluster ($\mathcal{L}_{2}$).
Furthermore, based on our analyses in Section~\ref{sec:analysis}, we employ a distance loss to increase the mean inter-group distances over the mean intra-group distances, enhancing the separability of the group ($\mathcal{L}_{3}$).
Specifically, each $c-$th cluster has a learnable centroid vector, denoted by $\mathbf{t}_{c} \in \mathbb{R}^{d}$, and the cluster-assignment score of node $v_{i}$ to the $c-$th cluster, denoted by $p_{ic}$, is defined as: $p_{ic} = \frac{\exp((\mathbf{x}_{i}^{*})^{T}\mathbf{t}_{c})}{\sum_{k\in [C]} \exp((\mathbf{x}_{k}^{*})^{T}\mathbf{t}_{c})}$, where $C$ is the number of clusters.
Then, each loss is defined formally as follows:
\begin{align}
    \mathcal{L}_{1}&=-\frac{1}{\vert \mathcal{V}\vert}\sum_{v_{i}\in \mathcal{V}}\sum_{c \in [C]}p_{ic} \log{p_{ic}}, \label{eq:clus1}\\
    \mathcal{L}_{2}&=\sum_{c \in [C]}\bar{p}_{c}\log{\bar{p}_{c}}, \text{ where }
    \bar{p}_{c}=\frac{1}{\vert \mathcal{V}\vert} \sum_{v_{i}\in \mathcal{V}}p_{ic}, \label{eq:clus2}\\
    \mathcal{L}_{3}&= \exp(\sum_{v_{i},v_{j} \in \mathcal{V}^{+}} \frac{d( \mathbf{x}^{*}_{i} , \mathbf{x}^{*}_{j})}{\vert \mathcal{V^{+}}\vert} - \sum_{v_{k},v_{\ell} \in 
    \mathcal{V}^{-}} \frac{d(\mathbf{x}^{*}_{k} , \mathbf{x}^{*}_{\ell})}{\vert \mathcal{V}^{-}\vert}), \label{eq:clus3}
\end{align}

where $e$ is an exponential function, $d(\mathbf{x},\mathbf{y}) = \lVert \mathbf{x} - \mathbf{y}\rVert_{2}$, $\mathcal{V}^{+} = \{\{v_{i},v_{j}\} : \mathcal{T}(v_{i}) = \mathcal{T}(v_{j}), \{v_{i},v_{j}\} \in \binom{\mathcal{V}}{2}\}$, $\mathcal{T}(v_{i}) = \argmax_{c\in [C]}p_{ic}$, and $\mathcal{V}^{-} = \binom{V}{2}\setminus \mathcal{V}^{+}$.
The final clustering loss is defined as 
$\mathcal{L}_{clus} = \mathcal{L}_{1} + \mathcal{L}_{2} + \lambda\mathcal{L}_{3}$, where $\lambda$ is a loss coefficient for the distance loss term.
All parameters (i.e., $c_{0},c_{1},c_{2}$ and $\mathbf{t}_{c},\forall c\in [C]$) are optimized using gradient descent aiming to minimize $\mathcal{L}_{clus}$.
After cluster training, using optimized weights, which are denoted as $\alpha^{*}_{0},\alpha^{*}_{1},$ and $\alpha^{*}_{2}$, we derive intermediate embeddings $\mathbf{H}$ as:
\begin{equation}\label{eq:intermediateemb}
\mathbf{H} = (\alpha^{*}_{0}\mathbf{I} + \alpha^{*}_{1}\tilde{\mathbf{A}} + \alpha_{2}^{*}\tilde{\mathbf{A}^{2}})\mathbf{X} \in \mathbb{R}^{\vert \mathcal{V} \vert \times d}.   
\end{equation}

\subsection{Step 2. Improving latent-class separability}\label{subsec:step2}

Recall that our adaptive graph convolution model (Eq.~\eqref{eq:adaptgc}) is designed to learn the adequate degree of graph convolution usage. 
However, due to its simplicity (i.e., limited expressiveness), it may not be optimal for producing embeddings that maximize latent-class (i.e., cluster) separability.
To enhance latent-class separability, we refine the intermediate node embeddings $\mathbf{H}$ (Eq.~\eqref{eq:intermediateemb}) obtained from Step 1 (Section~\ref{subsec:step1}) using a refinement model.

\noindent\textbf{Refinement model.}
As a refinement model that transforms the intermediate embeddings $\mathbf{H}$ into the final embeddings $\mathbf{Z}$, we use a feed-forward neural network $f_{\theta}$.
Moreover, a skip connection technique is incorporated to preserve the essential information encoded in $\mathbf{H}$.
Consequently, the final node embeddings $\mathbf{Z}$ are defined as
$\mathbf{Z}=f_{\theta}(\mathbf{H}) + \mathbf{H}$.
During refinement, the intermediate embeddings $\mathbf{H}$ remain fixed, and only the parameters $\theta$ of the refinement model are optimized.

\noindent\textbf{Training for refinement.}
Given intermediate node embeddings $\mathbf{H}$, we enhance their cohesion within each latent class (i.e., each cluster) while ensuring distinctiveness between different latent classes.
Specifically, we enhance the similarity of embeddings for nearest neighbors in the intermediate embedding space (i.e., $\mathbf{H}$ space) since they are likely to belong to the same latent class due to the latent-class separability in the $\mathbf{H}$ space.
Formally, we denote a set of $N$ nearest neighbors of each node $v_{i}$ in the $\mathbf{H}$ space as $\mathcal{V}'_{i}$.
In addition, the set of nearest node pairs is denoted as $\mathcal{V}_{+}' = \{\{v_{i},v_{j}\} : v_{j} \in \mathcal{V}'_{i},v_{i}\in \mathcal{V}\}$ and a set of negative pairs is denoted as $\mathcal{V}_{-}' =\binom{\mathcal{V}}{2} \setminus \mathcal{V}^{'}_{+}$.
We train $f_{\theta}$ to decrease the distance between node pairs in $\mathcal{V}'_{+}$ and increase the distance between node pairs in $\mathcal{V}'_{-}$.

As the objective function, we employ an exponential function, resulting in the overall loss function being a geometric average.
We provide a detailed description of the advantage of using this loss function in Section D.3 of the supplementary material.
Formally, the overall loss function for refinement is defined as follows:
\begin{equation}\label{eq:distanceloss}
    \mathcal{L}_{dist}= \exp(\sum_{v_{i},v_{j} \in \mathcal{V}'_{+}} \frac{d( \mathbf{z}_{i} , \mathbf{z}_{j})}{\tau\vert \mathcal{V}'_{+}\vert} - \sum_{v_{k},v_{\ell} \in 
    \mathcal{V}'_{-}} \frac{d(\mathbf{z}_{k} , \mathbf{z}_{\ell})}{\tau\vert \mathcal{V}'_{-}\vert}),
\end{equation}
where $d(\mathbf{x},\mathbf{y}) = \lVert \mathbf{x} - \mathbf{y}\rVert_{2}$ and $\tau$ is a temperature hyperparameter.
The parameters of the refinement model are trained using gradient descent aiming to minimize $\mathcal{L}_{dist}$ (Eq.~\eqref{eq:distanceloss}).
We provide a complexity analysis of \method in Section D.4 of the supplementary material.


\section{Experiments}
\label{sec:experiment}

In this section, we evaluate the effectiveness of \method in several downstream tasks.
To this end, we answer the following four research questions:
\begin{itemize}[leftmargin=*]
    \item {\textbf{RQ1.} How effective are the node embeddings obtained by \method in the node classification task?}
    \item {\textbf{RQ2.} How effective are the node embeddings obtained by \method in the clustering task?} 
    \item {\textbf{RQ3.} Does \method achieve its design goal?} 
    \item {\textbf{RQ4.} Are all the key components of \method essential for achieving high performance?}
\end{itemize}

\subsection{Experimental settings}

\begin{table*}[t]
\setlength{\tabcolsep}{1.5pt}
\small
\centering
\scalebox{0.85}{
\renewcommand{\arraystretch}{1.0}
\begin{tabular}{c | c |c c c c c c c c c c c c c c | c}
    \toprule
        & Datasets & Squirrel & Actor & Wis. & Cornell & Texas & Cham. & Penn94 & Flickr & Cora & Citeseer & Pubmed & Photo & Comp. & Arxiv & A.R.  \\

        \midrule 

        & Homophily & 0.217 & 0.220 & 0.155 & 0.111 & 0.057 & 0.247 & 0.483 & 0.322 & 0.825 & 0.717 & 0.792 & 0.849 & 0.802 & 0.635 & - \\
        
    \midrule 
    \midrule 

    & Naive $\mathbf{X}$ & 34.0 {\std (1.9)} & 35.4 {\std (0.9)} & 79.0 {\std (5.0)} & 71.1 {\std (3.4)} & 75.1 {\std (4.9)} & 49.8 {\std (1.3)} & 73.7 {\std (0.4)} & 49.9 {\std (0.1)} 
    & 58.5 {\std (0.7)} & 59.0 {\std (1.8)} & 72.8 {\std (0.1)} & 88.5 {\std (0.5)} & 81.8 {\std (0.6)}
    & 58.9 {\std (0.1)} & 14.0 \\

    \midrule
    
    \multirow{4}{*}{\rotatebox[origin=c]{90}{\textbf{CL}}}
     
     & DGI & 42.2 {\std (1.2)} & 29.8 {\std (1.5)} & 58.6 {\std (6.0)} & 48.9 {\std (6.8)} & 65.9 {\std (4.6)} & 59.8 {\std (2.5)} 
     & 67.6 {\std (0.8)} & 51.8 {\std (0.3)}
     & 82.4 {\std (0.6)} & 71.4 {\std (1.4)} & 79.4 {\std (0.8)} & 92.8 {\std (0.6)} & 87.5 {\std (0.5)} 
     & 71.2 {\std (0.3)} & 11.7 \\

     & GraphCL & 49.1 {\std (1.0)} & 30.7 {\std (0.6)} & 60.6 {\std (6.6)} & 42.4 {\std (7.8)} & 65.0 {\std (4.8)} & 64.5 {\std (2.6)}
     & O.O.M. & O.O.M.
     & 82.4 {\std (0.8)} & 71.3 {\std (0.9)} & \best 83.4 {\std (0.5)} & 93.0 {\std (0.5)} & \best 90.3 {\std (0.3)} 
     & O.O.M. & 10.3 \\

     & MVGRL & 49.1 {\std (1.0)} & 30.0 {\std (1.4)} & 69.7 {\std (4.6)} & 47.8 {\std (6.7)} & 69.7 {\std (4.6)} & 57.1 {\std (2.2)}
     & O.O.M. & 51.6 {\std (0.3)}
     & 82.0 {\std (1.2)} & 72.1 {\std (0.8)} & 78.8 {\std (0.9)} & {93.0 {\std (0.4)}} & {88.0 {\std (0.5)}} 
     & O.O.M. & 12.1 \\

     & BGRL & 55.0 {\std (0.6)} & 30.9 {\std (1.2)} & 59.4 {\std (6.2)} & 49.2 {\std (5.0)} & 64.3 {\std (3.8)} & 66.8 {\std (1.8)}
     & 74.2 {\std (0.7)} & \secb 53.1 {\std (0.1)}
     & 82.1 {\std (1.2)} & 71.2 {\std (0.6)} 
     & \secb 81.0 {\std (0.1)} & 93.2 {\std (0.3)} & 90.0 {\std (0.1)} 
     & \secb 71.8 {\std (0.2)} & 8.4 \\

     \midrule 

     \multirow{3}{*}{\rotatebox[origin=c]{90}{\textbf{Gen}}}

     & GAE & 47.8 {\std (1.7)} & 30.3 {\std (0.8)} & {54.3} {\std (7.0)} & 43.5 {\std (1.1)} & 67.0 {\std (5.5)} & 57.9 {\std (2.1)}
     & O.O.M. & O.O.M.
     & 81.9 {\std (1.0)} & 69.1 {\std (1.3)} & 76.6 {\std (1.6)} 
     & {89.0 {\std (0.3)}} & {93.0 {\std (0.3)}} 
     & O.O.M. & 14.0 \\

     & GraphMAE & 45.8 {\std (1.0)} & 29.8 {\std (1.0)} & 59.6 {\std (6.1)} & 50.8 {\std (8.3)} & 65.9 {\std (7.6)} & 66.2 {\std (2.4)} 
     & 59.1 {\std (0.4)} & 51.7 {\std (0.2)}
     & \secb 83.5 {\std (0.9)} &  73.0 {\std (1.0)} & 80.8 {\std (1.0)} & 92.6 {\std (0.3)} & 88.8 {\std (0.6)} 
     & 71.3 {\std (0.3)} & 9.7 \\

     & MaskGAE & 48.0  {\std (1.0)} & 29.0 {\std (1.4)} & 60.4 {\std (0.0)}  & 46.8 {\std (7.4)} & 70.5 {\std (5.6)} & 64.6 {\std (1.4)} 
     & 63.5 {\std (0.5)} & \best 53.2 {\std (0.2)}
     & 82.6 {\std (0.9)} & 71.4 {\std (0.7)} & 80.4 {\std (1.8)} & 93.8 {\std (0.3)} & 89.6 {\std (0.8)} 
     & 70.7 {\std (0.1)} & 9.5 \\

     \midrule 

     \multirow{7}{*}{\rotatebox[origin=c]{90}{\textbf{Non-homophilic}}}

     & HGRL & 46.3 {\std (2.1)} & 37.3 {\std (1.0)} & 82.2 {\std (3.1)} & 74.4 {\std (4.9)} & \secb 83.4 {\std (5.8)} & 61.1 {\std (1.9)} 
     & O.O.M. & O.O.M.
     & 80.2 {\std (0.7)} & 70.7 {\std (1.1)} & 78.8 {\std (0.8)} & 92.8 {\std (0.1)} & 88.1 {\std (0.3)} 
     & O.O.M. & 10.6 \\

     & DSSL & 53.5 {\std (1.4)} & 29.2 {\std (1.1)} & 60.0 {\std (5.3)} & 45.1 {\std (7.6)} & 64.1 {\std (4.5)} & 68.5 {\std (2.1)} 
     & O.O.M. & O.O.M.
     & 82.0 {\std (0.8)} & 66.0 {\std (1.3)} & 77.3 {\std (1.4)} & 92.8 {\std (0.3)} & 89.7 {\std (0.3)} 
     & O.O.M. & 13.0 \\

     & GREET & 49.2 {\std (2.0)} & \secb 37.6 {\std (1.3)} &  84.1 {\std (4.3)} & \secb 76.8 {\std (4.4)} & 81.1 {\std (5.7)} & 62.8 {\std (1.3)} 
     & O.O.M. & O.O.M.
     & 83.3 {\std (0.5)} & 72.5 {\std (0.8)} & 79.2 {\std (0.7)} & 92.8 {\std (0.3)} & 88.3 {\std (0.3)} 
     & O.O.M. & 7.8 \\

     & NeCo & 45.3 {\std (1.3)} & 30.5 {\std (1.1)} & 56.4 {\std (5.4)} & 54.3 {\std (6.2)} & 63.5 {\std (5.0)} & 58.3 {\std (2.3)} 
     & O.O.M. & O.O.M.
     & 82.3 {\std (0.8)} & 68.5 {\std (1.2)} & 80.7 {\std (0.7)} & 92.4 {\std (0.2)} & 89.1 {\std (0.3)} 
     & O.O.M. & 13.1 \\

     & HLCL & 41.3 {\std (1.7)} & 28.7 {\std (1.2)} & 60.2 {\std (4.4)} & 44.5 {\std (7.2)} & 58.4 {\std (3.7)} & 53.1 {\std (2.4)} 
     & O.O.M. & O.O.M.
     & 81.1 {\std (1.2)} & 70.3 {\std (0.7)} & 80.2 {\std (1.3)} & 91.5 {\std (0.3)} & 87.3 {\std (0.3)} 
     & O.O.M. & 15.5 \\

     & PolyGCL & 56.7 {\std (1.5)} & 35.0 {\std (1.0)} & 83.3 {\std (3.1)} & 74.8 {\std (6.2)} & 80.5 {\std (6.3)} & 70.7 {\std (2.1)} 
     & O.O.M. & O.O.M.
     & 82.6 {\std (1.2)} & 71.9 {\std (0.7)} & 78.8 {\std (1.0)} & 91.3 {\std (0.4)} & 87.4 {\std (0.6)} 
     & O.O.M. & 9.1 \\

     & HeterGCL & 42.1 {\std (1.1)} & 36.5 {\std (1.4)} & {82.4} {\std (3.6)} & {71.6} {\std (4.1)} & 78.9 {\std (5.6)} & 58.9 {\std (1.7)} 
     & O.O.M. & O.O.M.
     & 82.5 {\std (1.0)} & 71.9 {\std (1.1)} & 81.5 {\std (0.6)} & 93.0 {\std (0.3)} & 87.6 {\std (0.6)} 
     & O.O.M. & 9.2 \\

     \midrule

    \multirow{4}{*}{\rotatebox[origin=c]{90}{\textbf{Variants}}}

    & w/o Step 1 & 46.6 {\std (1.4)} & 34.2 {\std (0.8)} & 64.3 {\std (7.3)} & 54.9 {\std (6.4)} & 63.2 {\std (5.6)} & 52.1 {\std (2.0)} 
    & 74.3 {\std (0.5)} & 50.8 {\std (0.1)} & 83.5 {\std (0.2)} & 73.3 {\std (0.6)} & 79.4 {\std (0.5)} & 93.9 {\std (0.4)} & 90.1 {\std (0.3)} 
    & 71.7 {\std (0.2)} & 8.3 \\

    & w/o Step 2 & 56.6 {\std (1.7)} & 36.2 {\std (0.9)} & 79.0 {\std (5.2)} & 70.8 {\std (3.8)} & 75.1 {\std (5.0)} & 70.7 {\std (1.6)} 
    & \best 75.0 {\std (0.3)} & 51.0 {\std (0.3)} & 78.9 {\std (0.8)} & 67.4 {\std (0.7)} & 77.0 {\std (0.2)} & 94.0 {\std (0.3)} & 88.9 {\std (0.3)} 
    & 71.7 {\std (0.2)} & 8.6 \\

    & w/o SK & \secb 64.8 {\std (0.8)} & 34.4 {\std (0.7)} & \secb 86.7 {\std (3.6)} & 76.2 {\std (4.0)} & 77.8 {\std (4.2)} & \secb 72.4 {\std (2.0)} 
    & 71.6 {\std (0.4)} & 50.4 {\std (0.3)} & 82.7 {\std (0.7)} & \secb 73.4 {\std (0.7)} & 79.3 {\std (0.5)} & 93.9 {\std (0.4)} & 89.9 {\std (0.4)} 
    & 71.2 {\std (0.6)} & \secb 5.5 \\

    & w/o Exp & 64.6 {\std (1.0)} & 35.2 {\std (1.1)} &  85.9 {\std (2.7)} & 75.4 {\std (4.8)} & 78.1 {\std (4.4)} & 72.2 {\std (1.6)} 
    & 73.9 {\std (0.3)} & 50.7 {\std (0.1)} & 81.6 {\std (0.8)} & 72.1 {\std (0.8)} & 78.9 {\std (0.4)} & \secb 94.1 {\std (0.3)} & 89.7 {\std (0.4)} 
    & 71.8 {\std (0.2)} & 6.1 \\

    \midrule
    \midrule
    
    & \method & \best 65.2 {\std (0.7)} & \best 38.2 {\std (0.8)} & \best 87.6 {\std (4.1)} & \best 78.1 {\std (0.4)} & \best 84.6 {\std (5.0)} & \best 73.0 {\std (1.7)} 
    & \secb 74.6 {\std (0.5)} & 51.1 {\std (0.7)}
    & \best 83.8 {\std (0.3)} & \best 74.1 {\std (0.5)} & 79.6 {\std (0.6)} & \best 94.2 {\std (0.3)} & \secb 90.1 {\std (0.4)} 
    & \best 72.0 {\std (0.1)} & \best 2.4 \\
     
     \bottomrule
\end{tabular}
}
\caption{Node classification performance. Mean and standard deviation of test accuracy values ($\times$100) in the node classification task are reported.
The best and second-best performances are highlighted in green and yellow colors, respectively.
A.R. and O.O.M. denote average ranking and out-of-GPU memory, respectively.
0.0* indicates a value smaller than $10^{-4}$.
\method achieves the best average ranking among the 16 compared methods.
\label{tab:mainclassification}}
\end{table*}

\textbf{Datasets.}
We evaluate \method on 14 real-world graph datasets, comprising 8 non-homophilic graphs (edge homophily~\citep{zhu2020beyond} < 0.5) and 6 homophilic graphs (edge homophily > 0.5).
These datasets span seven diverse domains, with varying sizes (ranging from 183 to 167,343 nodes) and a wide range of homophily levels (edge homophily from 0.11 to 0.85).\footnote{For large-scale graphs with $\vert \mathcal{V}\vert > 4 \times 10^{4}$, we employ several scalable techniques, which are detailed in Section D.5 of the supplementary material.}
Additional dataset details and results for other versions of certain datasets (Chameleon and Squirrel) are in Section B and D.9 of~\cite{supplementary}, respectively.

\noindent\textbf{Baseline methods.} 
For comparison, we use input node features alone (denoted as \textbf{Naive} $\mathbf{X}$) and 14 unsupervised node representation learning baseline methods. 
These consist of 4 contrastive learning approaches (denoted as \textbf{CL}), 3 generative self-supervised learning approaches (denoted as \textbf{Gen}), and 7 non-homophilic graph unsupervised representation learning approaches (denoted as \textbf{Non-homophilic}).
Each category consists of the following:
\begin{itemize}[leftmargin=*]
    \item \textbf{CL:} DGI~\citep{velivckovic2018deep}, GraphCL~\citep{you2020graph}, MVGRL~\citep{hassani2020contrastive}, and BGRL~\citep{thakoor2021bootstrapped}.
    \item \textbf{Gen:} GAE~\citep{kipf2016variational}, GraphMAE~\citep{hou2022graphmae}, and MaskGAE~\citep{li2023s}.
    \item \textbf{Non-homophilic:} HGRL~\citep{chen2022towards}, DSSL~\citep{xiao2022decoupled}, GREET~\citep{liu2023beyond}, NeCo~\citep{he2023contrastive}, PolyGCL~\citep{chen2024polygcl}, HeterGCL~\citep{wang2024hetergcl}, and HLCL~\citep{yang2024graph}.
\end{itemize}


\noindent\textbf{Training and evaluation.}
To assess each method, we first obtain node embeddings using it, and then we use the embeddings as input features to perform downstream tasks, specifically node classification and clustering.
{Moreover, we provide qualitative analysis via embedding visualizations in Appendix D.14 of~\cite{supplementary}.}
{For \method, we choose the number of clusters to equal the number of node classes in the dataset; Appendix D.12 of~\cite{supplementary} further confirms that \method is not sensitive to this choice.}
For training/validation/test splits, we use fixed splits provided in PyG~\citep{fey2019fast}.
For datasets where fixed splits are not provided in PyG, we follow the node partitioning strategy proposed in the original works that introduced these datasets.
Further details for the splits are in Section B.3 of the supplementary material.
For each dataset–method pair, we run 10 trials, varying the model initialization and dataset splits. 
If a dataset provides only a single fixed split, we use the same split for all experiment trials.
We perform the hyperparameter tuning with the validation set and use the configurations that give the best validation performance for the evaluation.
Further details regarding the hyperparameter tuning are in Section C.2 of the supplementary material.

\begin{table*}[t]
\setlength{\tabcolsep}{1.5pt}
\small
\centering
\scalebox{0.85}{
\renewcommand{\arraystretch}{1.0}
\begin{tabular}{c | c |c c c c c c c c c c c c c c | c}
    \toprule
        & Datasets & Squirrel & Actor & Wis. & Cornell & Texas & Cham. & Penn94 & Flickr & Cora & Citeseer & Pubmed & Photo & Comp. & Arxiv & A.R.  \\

        \midrule 

        & Homophily & 0.217 & 0.220 & 0.155 & 0.111 & 0.057 & 0.0 & 0.483 & 0.322 & 0.825 & 0.717 & 0.792 & 0.849 & 0.802 & 0.635 & - \\
    \midrule 
    \midrule 

    & Naive $\mathbf{X}$ & 0.0 {\std (0.0)} & 5.4 {\std (0.4)} & 32.8 {\std (2.4)} & 9.2 {\std (0.8)} & 17.1 {\std (6.6)} & 9.1 {\std (2.6)} 
    & 1.8 {\std (0.9)} & 1.0 {\std (0.0*)} 
    & 18.6 {\std (3.4)} & 20.6 {\std (2.6)} & 31.0 {\std (0.0*)} & 13.6 {\std (1.4)} & 12.3 {\std (1.2)} 
    & 21.9 {\std (0.1)} & 12.9 \\

    \midrule
    
    \multirow{4}{*}{\rotatebox[origin=c]{90}{\textbf{CL}}}
     
     & DGI & 6.8 {\std (0.2)} & 0.1 {\std (0.0*)} & 14.3 {\std (2.3)} & 11.9 {\std (1.6)} & 18.3 {\std (2.0)} & 14.3 {\std (1.0)}
     & \secb 1.8 {\std (0.0*)} & 6.7 {\std (0.5)}
     & 54.4 {\std (1.1)} & 38.6 {\std (1.4)} & 16.5 {\std (1.4)} & 48.2 {\std (2.2)} & 43.2 {\std (1.7)} & 35.3 {\std (0.4)} & 11.3 \\

     & GraphCL & 6.6 {\std (0.2)} & 0.5 {\std (0.0)} & 13.1 {\std (2.2)} & 10.5 {\std (2.3)} & 16.1 {\std (2.3)} & 13.7 {\std (0.6)} 
         & O.O.M. & O.O.M. 
         & \secb 58.2 {\std (1.5)} & 43.3 {\std (0.9)} &  33.5 {\std (1.1)} & 53.7 {\std (2.8)} &  47.5 {\std (1.9)} & O.O.M. & 10.6 \\

     & MVGRL & 6.0 {\std (0.4)} & 0.1 {\std (0.0*)} & 13.5 {\std (1.9)} & 7.0 {\std (1.7)} & 22.3 {\std (1.1)} & 12.5 {\std (0.6)} 
     & O.O.M. & \secb 6.8 {\std (0.5)}
     & 58.1 {\std (1.2)} & 44.2 {\std (0.8)} & \best 33.8 {\std (1.0)} & {46.5 {\std (3.0)}} & {40.3 {\std (2.7)}} & O.O.M. & 10.9 \\

     & BGRL & 5.5 {\std (0.8)} & 1.0 {\std (0.4)} & 12.3 {\std (2.5)} & 10.0 {\std (1.2)} & 15.8 {\std (0.9)} & 14.7 {\std (0.5)} 
     & 1.4 {\std (0.3)} & \best 7.2 {\std (0.3)}
     & 55.0 {\std (3.2)} 
     & 43.2 {\std (1.0)} & \secb 32.5 {\std (1.2)} & 51.8 {\std (3.4)} & 40.0 {\std (1.0)} & 34.3 {\std (1.1)} & 11.3 \\

     \midrule 

     \multirow{3}{*}{\rotatebox[origin=c]{90}{\textbf{Gen}}}

     & GAE & 7.2 {\std (0.2)} & 1.5 {\std (0.1)} & 8.1 {\std (0.8)} & 11.6 {\std (0.8)} & 17.1 {\std (3.3)} & 14.4 {\std (0.8)} 
     & O.O.M. & O.O.M.
     & 45.5 {\std (4.9)} & 33.8 {\std (4.2)} & 28.5 {\std (1.1)} 
     & {45.1 {\std (1.0)}} & {38.0 {\std (0.5)}} & O.O.M. & 13.2 \\

     & GraphMAE & 6.8 {\std (0.3)} & 1.4 {\std (0.0)} & 16.7 {\std (1.1)} & 16.5 {\std (1.0)} & 19.7 {\std (2.0)} & 14.9 {\std (0.6)} 
     & 1.7 {\std (0.0*)} & 4.7 {\std (0.1)}
     & 56.5 {\std (1.2)} & \secb 44.3 {\std (0.9)} & 25.7 {\std (1.0)} & 53.4 {\std (4.1)} & 32.8 {\std (6.9)} & 22.2 {\std (1.2)} & 9.2 \\

     & MaskGAE & 7.2 {\std (0.7)} & 1.2 {\std (0.1)} & 10.9 {\std (2.1)}  & 7.0 {\std (1.1)} & 19.4 {\std (2.2)} & 14.7 {\std (0.5)} 
     & 0.7 {\std (0.1)} & 5.5 {\std (0.5)}
     & 55.5 {\std (2.8)} & 42.9 {\std (0.6)} & 14.3 {\std (5.4)} & 51.6 {\std (2.7)} & 33.4 {\std (4.1)} & 36.4 {\std (0.5)} & 11.8 \\

     \midrule 

     \multirow{7}{*}{\rotatebox[origin=c]{90}{\textbf{Non-homophilic}}}

     & HGRL & 8.2 {\std (0.1)} & 5.9 {\std (0.5)} & 36.2 {\std (1.6)} &  40.1 {\std (3.9)} &  38.5 {\std (3.3)} & \best 21.9 {\std (0.4)} 
     & O.O.M. & O.O.M.
     & 51.3 {\std (1.8)} & 42.4 {\std (1.5)} & 25.2 {\std (2.7)} & 50.2 {\std (2.0)} & 46.8 {\std (2.0)} & O.O.M. & 8.5 \\

     & DSSL & 6.6 {\std (0.7)} & 1.8 {\std (0.2)} & 12.1 {\std (1.9)} & 10.5 {\std (1.4)} & 16.6 {\std (2.2)} & 15.2 {\std (0.4)} 
     & O.O.M.& O.O.M.
     & 54.3 {\std (2.6)} & 33.7 {\std (1.7)} & 15.6 {\std (4.4)} & 69.9 {\std (0.9)} & 47.2 {\std (2.8)} & O.O.M. & 12.2 \\

     & GREET & 6.2 {\std (0.3)} & \secb 5.9 {\std (0.5)} & 43.7 {\std (2.8)} & 40.0 {\std (2.0)} & 37.5 {\std (2.0)} & \secb 21.6 {\std (0.5)} 
     & O.O.M. & O.O.M.
     & 57.3 {\std (1.7)} & 43.4 {\std (1.8)} & 24.8 {\std (0.6)} & 55.5 {\std (1.4)} & 41.1 {\std (1.3)} & O.O.M. & 8.1 \\

     & NeCo & 5.2 {\std (0.4)} & 0.3 {\std (0.0*)} & 12.7 {\std (1.1)} & 17.9 {\std (0.3)} & 17.9 {\std (1.4)} & 13.2 {\std (1.6)} 
     & O.O.M. & O.O.M.
     & 34.0 {\std (1.0)} & 16.9 {\std (2.8)} & 32.1 {\std (1.4)} & 39.1 {\std (1.6)} & 32.5 {\std (1.8)} & O.O.M. & 14.7 \\

     & HLCL & 6.3 {\std (0.3)} & 0.6 {\std (0.2)} & 12.8 {\std (0.2)} & 10.4 {\std (4.8)} & 17.9 {\std (2.2)} & 14.4 {\std (1.2)} 
     & O.O.M. & O.O.M.
     & 55.6 {\std (1.7)} & 40.4 {\std (0.7)} & 16.8 {\std (0.2)} & 55.7 {\std (2.9)} & 46.1 {\std (0.5)} & O.O.M. & 12.4 \\

     & PolyGCL & 7.8 {\std (0.4)} & 4.3 {\std (0.7)} & \secb 44.8 {\std (3.0)} & 19.8 {\std (5.4)} & 34.9 {\std (3.3)} & 19.4 {\std (0.5)} 
     & O.O.M. & O.O.M.
     & 35.8 {\std (3.6)} & 12.4 {\std (2.8)} & 28.0 {\std (0.9)} & 42.2 {\std (1.5)} & 34.0 {\std (3.9)} & O.O.M.
     & 10.7 \\

     & HeterGCL & 5.4 {\std (0.1)} & 3.3 {\std (1.1)} & 27.0 {\std (4.8)} & 18.1 {\std (2.2)} & 23.1 {\std (1.2)} & 15.3 {\std (4.2)} 
     & O.O.M. & O.O.M.
     & 51.0 {\std (2.2)} & 41.5 {\std (1.1)} & 32.3 {\std (0.3)} 
     & 56.7 {\std (2.6)} & 47.4 {\std (1.2)} & O.O.M.
     & 9.6 \\

     \midrule

     \multirow{4}{*}{\rotatebox[origin=c]{90}{\textbf{Variants}}}

     & w/o Step 1 & 4.4 {\std (0.1)} & 0.8 {\std (0.1)} & 17.4 {\std (2.9)} & 12.4 {\std (0.5)} & 14.3 {\std (1.7)} & 14.1 {\std (1.0)} 
     & 0.9 {\std (0.0*)} & 2.4 {\std (0.0*)} & 55.9 {\std (0.9)} & 43.0 {\std (1.9)} & 31.7 {\std (0.6)} & \secb 71.0 {\std (0.9)} & 53.2 {\std (0.2)} 
    & 38.2 {\std (0.9)} & 9.5 \\

    & w/o Step 2 & 5.6 {\std (0.5)} & 5.4 {\std (0.1)} & 43.4 {\std (2.8)} & 35.1 {\std (3.3)} & 34.0 {\std (3.6)} & 9.5 {\std (1.2)} 
    & 0.8 {\std (0.0*)} & 2.3 {\std (0.2)} & 48.4 {\std (0.1)} & 40.9 {\std (1.8)} & 30.6 {\std (0.1)} & 66.2 {\std (0.9)} & 47.4 {\std (1.5)} 
    & 37.1 {\std (0.1)} & 9.3 \\

    & w/o SK & 8.3 {\std (0.3)} & 5.7 {\std (0.4)} & 47.7 {\std (0.5)} & \best 51.0 {\std (0.1)} & 39.5 {\std (2.0)} & 17.3 {\std (1.4)} 
    & 1.4 {\std (0.1)} & 2.0 {\std (0.1)} & 52.6 {\std (0.9)} & 44.6 {\std (0.1)} & 26.0 {\std (2.1)} & 71.1 {\std (0.1)} & \secb 53.5 {\std (0.3)} 
    & 39.6 {\std (0.3)} & 4.9 \secb \\

    & w/o Exp & \secb 8.8 {\std (0.4)} & 1.6 {\std (0.6)} & 45.6 {\std (1.3)} & 43.5 {\std (0.4)} & \secb 40.4 {\std (0.3)} & 10.4 {\std (0.1)} 
    & 1.2 {\std (0.0*)} & 1.9 {\std (0.1)} & 56.1 {\std (0.6)} & 42.5 {\std (0.9)} & 29.1 {\std (2.5)} & 70.9 {\std (0.3)} &  53.2 {\std (0.5)} 
    & \secb 40.0 {\std (0.1)} & 6.4 \\

     \midrule
     \midrule

    & \method & \best 9.5 {\std (0.2)} & \best 6.7 {\std (0.1)} & \best 50.1 {\std (0.9)} & \secb 43.8 {\std (0.6)} & \best 40.8 {\std (2.0)} & 16.8 {\std (1.4)} 
    & \best 1.9 {\std (0.0*)} & 2.2 {\std (0.0*)}
    & \best 59.0 {\std (0.3)} & \best 46.1 {\std (0.3)} & 30.7 {\std (0.4)} & \best 71.3 {\std (0.2)} & \best 54.4 {\std (0.3)} 
    & \best 40.3 {\std (0.1)} & \best 2.4 \\
     
     \bottomrule
\end{tabular}
}
\caption{{Clustering performance.} Mean and standard deviation of NMI (Normalized Mutual Information) values between the estimated clusters and the ground-truth clusters ($\times$100) are reported.
The best and second-best performances are highlighted in green and yellow.
A.R. and O.O.M. denote average ranking and out-of-GPU memory, respectively.
0.0* indicates a value smaller than $10^{-4}$.
\method achieves the best average ranking among the 16 compared methods.
\label{tab:mainclustering}}
\end{table*}

\subsection{RQ1. Node classification results}\label{subsec:classification}

In this section, we evaluate the effectiveness of each method in the node classification task.

\noindent\textbf{Settings.}
We train an MLP classifier using node embedding obtained by each method as input features and cross-entropy as the loss function. 
The MLP classifier is then used to measure test accuracy on node classification tasks.
{Further results using (1) linear classifiers and (2) noisy features are in Appendices D.7 and D.13 of~\cite{supplementary}, respectively.}


\noindent\textbf{Results.}
As shown in Table~\ref{tab:mainclassification}, \method achieves the best average ranking among 16 methods, demonstrating its effectiveness in learning representations for node classification.
Notably, \method performs overall best on graphs with diverse graph-level homophily, achieving the best performance on both the least homophilic graph (Texas) and the most homophilic graph (Photo). 

\subsection{RQ2. Clustering results}\label{subsec:clustering}

In this section, we evaluate the effectiveness of each method in the node clustering task.

\noindent\textbf{Settings.}
For each unsupervised node representation learning method, we cluster the learned embeddings using $K$-Means, setting $K$ to the number of unique labels in the dataset. 
We then compute the Normalized Mutual Information (NMI) between the resulting clusters and the ground-truth node labels, 
and regard this NMI score as the measure of clustering performance, following~\citet{wang2024hetergcl}. 
Additional results using the Adjusted Rand Index (ARI) metric are presented in Section D.8 of the supplementary material.

\noindent\textbf{Results.}
As shown in Table~\ref{tab:mainclustering}, \method achieves the best average ranking among 16 methods, demonstrating that \method's effectiveness extends beyond node classification to clustering. 
Consistent with its performance in node classification, \method achieves the best results on both the most and least homophilic graphs, further confirming its adaptability across diverse homophily levels in clustering tasks.

\begin{figure}[!t]  
  \centering

  \begin{minipage}[t]{\linewidth}
    \includegraphics[width=0.9\textwidth]{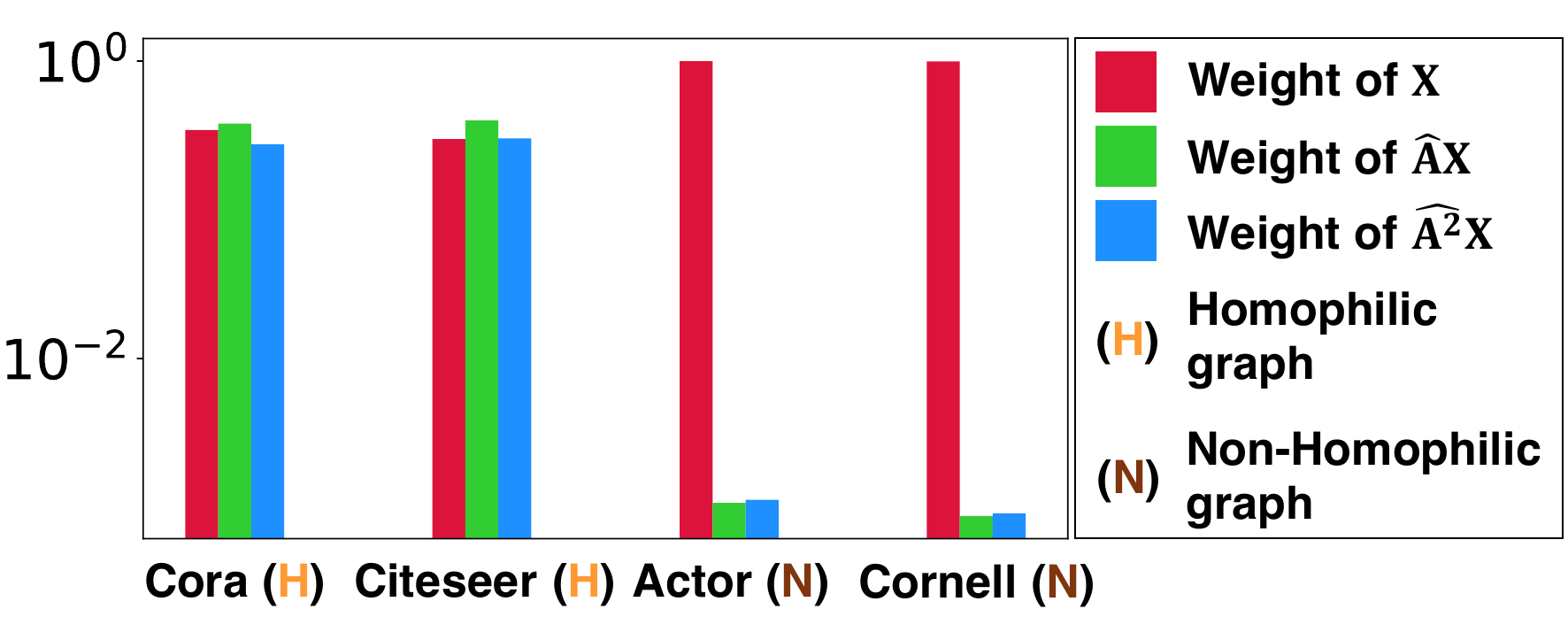}
    
    \small
    {(a)} Learned degree of graph convolution usage in Step 1 of \method (i.e., weights within the intermediate embeddings).
  \end{minipage}

  \begin{minipage}[t]{\linewidth}
  \small
    \includegraphics[width=0.9\textwidth]{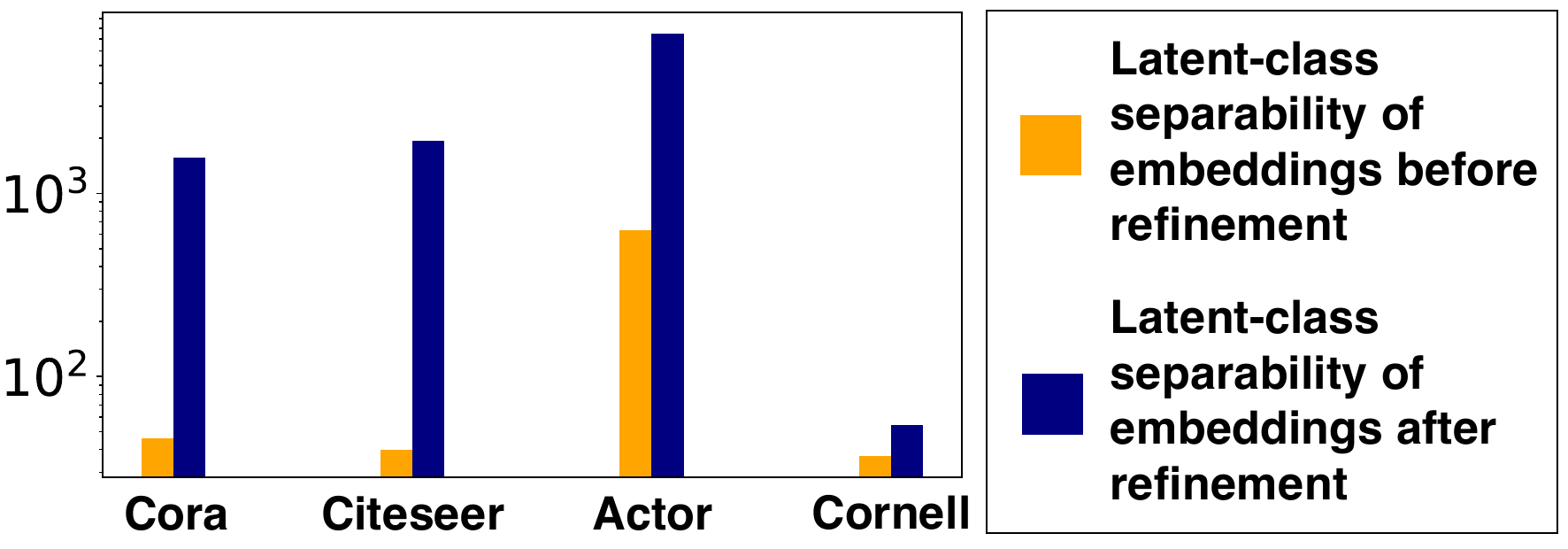}
    
    {(b)} Latent-class separability of the intermediate embeddings (before refinement) and the final embeddings (after refinement).
  \end{minipage}

  \caption{{Achievement of the design goals of  \method.}
  {(a)} demonstrates that \method adaptively learns the degree of graph convolution usage according to the level of graph homophily.
  {(b)} demonstrates that the refinement step of \method enhances the latent-class separability of embeddings.
  }
  \label{fig:sanitycheck}
\end{figure}

\subsection{RQ3. Achievement of design goals}

In this section, we evaluate whether \method meets its design objectives on real-world graphs.

\noindent\textbf{Settings.}
We aim to evaluate whether \method (1) adaptively learns the degree of graph convolution usage (Section~\ref{subsec:step1}) and (2) enhances the latent-class separability of embeddings through refinement (Section~\ref{subsec:step2}).
For the first goal, we analyze the learned weights in the intermediate embeddings $\mathbf{H}$ (Eq.~\eqref{eq:intermediateemb}): the weight of $\mathbf{X}$ ($\alpha^{*}_{0}$), the weight of $\mathbf{\hat{A}X}$ ($\alpha^{*}_{1}$), and the weight of $\hat{\mathbf{A^{2}}}\mathbf{X}$ ($\alpha^{*}_{2}$).
This analysis is conducted on both homophilic graphs (Cora and Citeseer) and non-homophilic graphs (Actor and Cornell) to verify whether each weight is adaptively learned based on the respective graph's homophily level.
For the second goal, we compare the latent-class separability of the intermediate embeddings (before refinement), and the final embeddings (after refinement) using the Calinski–Harabasz index, as in Section~\ref{subsec:motivation}.

\noindent\textbf{Results.}
First, as shown in Figure~\ref{fig:sanitycheck} (a), \method adaptively adjusts the degree of graph convolution based on graph homophily. 
Specifically, in homophilic graphs (Cora and Citeseer), \method assigns a significantly higher sum of weights to graph-convolution-related terms ($\mathbf{\hat{A}X}$ and $\mathbf{\hat{A^2}X}$) compared to the node features $\mathbf{X}$. 
In contrast, in non-homophilic graphs (Actor and Cornell), the weights assigned to graph-convolution-related terms are reduced to near zero.
Second, as shown in Figure~\ref{fig:sanitycheck} (b), \method significantly enhances the latent-class separability of the embeddings through its refinement step. 
These results demonstrate that \method successfully achieves its design objectives on real-world graphs with various homophily levels.

\subsection{RQ4. Ablation study}\label{subsec:ablation}

In this section, we assess whether each key component of \method is essential for achieving high performance.

\noindent\textbf{Settings.} 
We use 4 variants: (1) \method that does not use the specialized clustering scheme (w/o Step 1), (2) \method that does not perform the refinement step (w/o Step 2), (3) \method that does not use skip connection in the refiner (w/o SK), and (4) \method that does not use an exponential function for the loss (Eq.~\eqref{eq:distanceloss}) (w/o Exp).
Details on the variants are provided in Section C.3 of the supplementary material.
We evaluate each variant in both node classification and clustering, under the same setting of Section~\ref{subsec:classification} and \ref{subsec:clustering}, respectively.

\noindent\textbf{Results.} As shown in Tables~\ref{tab:mainclassification} and~\ref{tab:mainclustering}, 
\method outperforms its variants in 10 out of 14 datasets for both node classification and clustering tasks (see rows corresponding to \textbf{Variants}).
This result highlights the necessity of \method's key modules in obtaining high-quality node embeddings.

\section{Conclusion}
\label{sec:conclusion}
In this work, we investigate unsupervised node representation learning without the homophily assumption.
We begin by providing both theoretical and empirical analyses that highlight the necessity of adjusting the degree of graph convolution usage and suggesting an unsupervised method for this adjustment (Section~\ref{sec:analysis}).
Building on these insights, we introduce \method, a feature-centric approach that adaptively learns the adequate degree of graph convolution usage (Section~\ref{sec:method}).
Through extensive experiments including 15 baseline methods and 14 benchmark datasets, we demonstrate the effectiveness of \method in both node classification and clustering tasks (Section~\ref{sec:experiment}).
{As future work, we believe our approach can be applied to learning node representations on more diverse graph types, including heterogeneous~\cite{wang2022survey, zhao2021heterogeneous} and text-attributed graphs~\cite{yan2023comprehensive, kim2025hello}.
For reproducibility, code and datasets are in~\url{https://github.com/kswoo97/unsupervised-non-homophilic}.}

\section*{Ethical statement}
This work proposes an unsupervised technique for learning node representations. 
The experiments are conducted on publicly available benchmark datasets that have been widely used in prior work. We do not collect new human subject data, nor do we attempt to infer sensitive attributes.
While we believe our experiments do not involve any harmful or toxic content, applying our method in sensitive domains (e.g., social or financial networks) requires careful use to avoid potential privacy violations or unintended negative impacts.

\section*{Acknowledgements}
This work was partly supported by the National Research Foundation of Korea (NRF) grant funded by the Korea government (MSIT) (No. RS-2024-00406985, 40\%).
This work was partly supported by Institute of Information \& Communications Technology Planning \& Evaluation (IITP) grant funded by the Korea government (MSIT) (No. RS-2022-II220871, Development of AI Autonomy and Knowledge Enhancement for AI Agent Collaboration, 50\%)
(No. RS-2019-II190075, Artificial Intelligence Graduate School Program (KAIST), 10\%).

\bibliography{aaai2026}

@inproceedings{zhu2020beyond,
  title={Beyond homophily in graph neural networks: Current limitations and effective designs},
  author={Zhu, Jiong and Yan, Yujun and Zhao, Lingxiao and Heimann, Mark and Akoglu, Leman and Koutra, Danai},
  booktitle={NeurIPS},
  year={2020}
}

@inproceedings{velivckovic2018deep,
  title={Deep graph infomax},
  author={Veli{\v{c}}kovi{\'c}, Petar and Fedus, William and Hamilton, William L and Li{\`o}, Pietro and Bengio, Yoshua and Hjelm, R Devon},
  booktitle={ICLR},
  year={2019}
}

@inproceedings{you2020graph,
  title={Graph contrastive learning with augmentations},
  author={You, Yuning and Chen, Tianlong and Sui, Yongduo and Chen, Ting and Wang, Zhangyang and Shen, Yang},
  booktitle={NeurIPS},
  year={2020}
}

@inproceedings{hassani2020contrastive,
  title={Contrastive multi-view representation learning on graphs},
  author={Hassani, Kaveh and Khasahmadi, Amir Hosein},
  booktitle={ICML},
  year={2020}
}

@inproceedings{thakoor2021bootstrapped,
  title={Bootstrapped representation learning on graphs},
  author={Thakoor, Shantanu and Tallec, Corentin and Azar, Mohammad Gheshlaghi and Munos, R{\'e}mi and Veli{\v{c}}kovi{\'c}, Petar and Valko, Michal},
  booktitle={ICLR workshop on geometrical and topological representation learning},
  year={2021}
}

@inproceedings{hou2022graphmae,
  title={Graphmae: Self-supervised masked graph autoencoders},
  author={Hou, Zhenyu and Liu, Xiao and Cen, Yukuo and Dong, Yuxiao and Yang, Hongxia and Wang, Chunjie and Tang, Jie},
  booktitle={KDD},
  year={2022}
}

@inproceedings{kipf2016variational,
  title={Variational graph auto-encoders},
  author={Kipf, Thomas N and Welling, Max},
  booktitle={NeurIPS workshop on bayesian deep learning},
  year={2016}
}

@inproceedings{li2023s,
  title={What's Behind the Mask: Understanding Masked Graph Modeling for Graph Autoencoders},
  author={Li, Jintang and Wu, Ruofan and Sun, Wangbin and Chen, Liang and Tian, Sheng and Zhu, Liang and Meng, Changhua and Zheng, Zibin and Wang, Weiqiang},
  booktitle={KDD},
  year={2023}
}

@inproceedings{xiao2022decoupled,
  title={Decoupled self-supervised learning for graphs},
  author={Xiao, Teng and Chen, Zhengyu and Guo, Zhimeng and Zhuang, Zeyang and Wang, Suhang},
  booktitle={NeurIPS},
  year={2022}
}

@inproceedings{liu2023beyond,
  title={Beyond smoothing: Unsupervised graph representation learning with edge heterophily discriminating},
  author={Liu, Yixin and Zheng, Yizhen and Zhang, Daokun and Lee, Vincent CS and Pan, Shirui},
  booktitle={AAAI},
  year={2023}
}

@inproceedings{yang2024graph,
  title={Graph Contrastive Learning under Heterophily via Graph Filters},
  author={Yang, Wenhan and Mirzasoleiman, Baharan},
  booktitle={UAI}, 
  year = {2024}
}

@inproceedings{chen2022towards,
  title={Towards self-supervised learning on graphs with heterophily},
  author={Chen, Jingfan and Zhu, Guanghui and Qi, Yifan and Yuan, Chunfeng and Huang, Yihua},
  booktitle={CIKM},
  year={2022}
}

@inproceedings{he2023contrastive,
  title={Contrastive learning meets homophily: two birds with one stone},
  author={He, Dongxiao and Zhao, Jitao and Guo, Rui and Feng, Zhiyong and Jin, Di and Huang, Yuxiao and Wang, Zhen and Zhang, Weixiong},
  booktitle={ICML},
  year={2023}
}

@inproceedings{chen2024polygcl,
  title={PolyGCL: GRAPH CONTRASTIVE LEARNING via Learnable Spectral Polynomial Filters},
  author={Chen, Jingyu and Lei, Runlin and Wei, Zhewei},
  booktitle={ICLR},
  year={2024}
}

@inproceedings{wang2024hetergcl,
  title={HeterGCL: Graph Contrastive Learning Framework on Heterophilic Graph},
  author={Wang, Chenhao and Liu, Yong and Yang, Yan and Li, Wei},
  booktitle={IJCAI},
  year = {2024}
}

@inproceedings{fey2019fast,
  title={Fast graph representation learning with PyTorch Geometric},
  author={Fey, Matthias and Lenssen, Jan Eric},
  booktitle={ICLR workshop on representation learning on graphs and manifolds},
  year={2019}
}

@article{zheng2022graph,
  title={Graph neural networks for graphs with heterophily: A survey},
  author={Zheng, Xin and Wang, Yi and Liu, Yixin and Li, Ming and Zhang, Miao and Jin, Di and Yu, Philip S and Pan, Shirui},
  journal={arXiv preprint arXiv:2202.07082},
  year={2022}
}

@inproceedings{asano2019self,
  title={Self-labelling via simultaneous clustering and representation learning},
  author={Asano, Yuki Markus and Rupprecht, Christian and Vedaldi, Andrea},
  booktitle={ICLR},
  year={2020}
}

@inproceedings{caron2020unsupervised,
  title={Unsupervised learning of visual features by contrasting cluster assignments},
  author={Caron, Mathilde and Misra, Ishan and Mairal, Julien and Goyal, Priya and Bojanowski, Piotr and Joulin, Armand},
  booktitle={NeurIPS},
  year={2020}
}

@inproceedings{kipf2016semi,
  title={Semi-supervised classification with graph convolutional networks},
  author={Kipf, Thomas N and Welling, Max},
  booktitle={ICLR},
  year={2017}
}

@inproceedings{wu2019simplifying,
  title={Simplifying graph convolutional networks},
  author={Wu, Felix and Souza, Amauri and Zhang, Tianyi and Fifty, Christopher and Yu, Tao and Weinberger, Kilian},
  booktitle={ICML},
  year={2019}
}

@article{calinski1974dendrite,
  title={A dendrite method for cluster analysis},
  author={Cali{\'n}ski, Tadeusz and Harabasz, Jerzy},
  journal={Communications in Statistics-theory and Methods},
  volume={3},
  number={1},
  pages={1--27},
  year={1974},
  publisher={Taylor \& Francis}
}

@book{bishop2006pattern,
  title={Pattern recognition and machine learning},
  author={Bishop, Christopher M and Nasrabadi, Nasser M},
  volume={4},
  year={2006},
  publisher={Springer}
}

@inproceedings{khorasgani2022slic,
  title={Slic: Self-supervised learning with iterative clustering for human action videos},
  author={Khorasgani, Salar Hosseini and Chen, Yuxuan and Shkurti, Florian},
  booktitle={CVPR},
  year={2022}
}

@inproceedings{walawalkar2025videoclusternet,
  title={VideoClusterNet: Self-supervised and Adaptive Face Clustering for Videos},
  author={Walawalkar, Devesh and Garrido, Pablo},
  booktitle={ECCV},
  year={2024},
}

@inproceedings{zhao2023incremental,
  title={Incremental generalized category discovery},
  author={Zhao, Bingchen and Mac Aodha, Oisin},
  booktitle={ICCV},
  year={2023}
}

@inproceedings{ghasedi2017deep,
  title={Deep clustering via joint convolutional autoencoder embedding and relative entropy minimization},
  author={Ghasedi Dizaji, Kamran and Herandi, Amirhossein and Deng, Cheng and Cai, Weidong and Huang, Heng},
  booktitle={CVPR},
  year={2017}
}

@inproceedings{huang2020deep,
  title={Deep semantic clustering by partition confidence maximisation},
  author={Huang, Jiabo and Gong, Shaogang and Zhu, Xiatian},
  booktitle={CVPR},
  year={2020}
}

@inproceedings{guo2023architecture,
  title={Architecture matters: Uncovering implicit mechanisms in graph contrastive learning},
  author={Guo, Xiaojun and Wang, Yifei and Wei, Zeming and Wang, Yisen},
  booktitle={NeurIPS},
  year={2023}
}

@inproceedings{chien2020adaptive,
  title={Adaptive universal generalized pagerank graph neural network},
  author={Chien, Eli and Peng, Jianhao and Li, Pan and Milenkovic, Olgica},
  booktitle={ICLR},
  year={2021}
}

@inproceedings{grover2016node2vec,
  title={node2vec: Scalable feature learning for networks},
  author={Grover, Aditya and Leskovec, Jure},
  booktitle={KDD},
  year={2016}
}

@inproceedings{perozzi2014deepwalk,
  title={Deepwalk: Online learning of social representations},
  author={Perozzi, Bryan and Al-Rfou, Rami and Skiena, Steven},
  booktitle={KDD},
  year={2014}
}

@inproceedings{luan2022revisiting,
  title={Revisiting heterophily for graph neural networks},
  author={Luan, Sitao and Hua, Chenqing and Lu, Qincheng and Zhu, Jiaqi and Zhao, Mingde and Zhang, Shuyuan and Chang , Xiao-Wen and Precup, Doina},
  booktitle={NeurIPS},
  year={2022}
}

@misc{supplementary,
  title        = {Supplementary materials for this work},
  howpublished = {\url{https://github.com/kswoo97/unsupervised-non-homophilic}},
  author = {Kim, Sunwoo and Lee, Soo Yong and Kim, Kyungho and Hwang, Hyunjin and Yoo, Jaemin and Shin, Kijung},
  year = {2026}
}

@inproceedings{yu2024lg,
  title={LG-GNN: local-global adaptive graph neural network for modeling both homophily and heterophily},
  author={Yu, Zhizhi and Feng, Bin and He, Dongxiao and Wang, Zizhen and Huang, Yuxiao and Feng, Zhiyong},
  booktitle={IJCAI},
  year={2024}
}

@inproceedings{gasteiger2018predict,
  title={Predict then propagate: Graph neural networks meet personalized pagerank},
  author={Gasteiger, Johannes and Bojchevski, Aleksandar and G{\"u}nnemann, Stephan},
  booktitle={ICLR},
  year={2019}
}

@inproceedings{lee2024feature,
  title={Feature distribution on graph topology mediates the effect of graph convolution: Homophily perspective},
  author={Lee, Soo Yong and Kim, Sunwoo and Bu, Fanchen and Yoo, Jaemin and Tang, Jiliang and Shin, Kijung},
  booktitle={ICML},
  year={2024}
}

@inproceedings{kim2024hypeboy,
  title={Hypeboy: Generative self-supervised representation learning on hypergraphs},
  author={Kim, Sunwoo and Kang, Shinhwan and Bu, Fanchen and Lee, Soo Yong and Yoo, Jaemin and Shin, Kijung},
  booktitle={ICLR},
  year={2024}
}

@inproceedings{yan2023comprehensive,
  title={A comprehensive study on text-attributed graphs: Benchmarking and rethinking},
  author={Yan, Hao and Li, Chaozhuo and Long, Ruosong and Yan, Chao and Zhao, Jianan and Zhuang, Wenwen and Yin, Jun and Zhang, Peiyan and Han, Weihao and Sun, Hao and others},
  booktitle={NeurIPS},
  year={2023}
}

@inproceedings{hou2023graphmae2,
  title={Graphmae2: A decoding-enhanced masked self-supervised graph learner},
  author={Hou, Zhenyu and He, Yufei and Cen, Yukuo and Liu, Xiao and Dong, Yuxiao and Kharlamov, Evgeny and Tang, Jie},
  booktitle={WWW},
  year={2023}
}

@inproceedings{ju2022multi,
  title={Multi-task self-supervised graph neural networks enable stronger task generalization},
  author={Ju, Mingxuan and Zhao, Tong and Wen, Qianlong and Yu, Wenhao and Shah, Neil and Ye, Yanfang and Zhang, Chuxu},
  booktitle={ICLR},
  year={2023}
}

@article{wang2022survey,
  title={A survey on heterogeneous graph embedding: methods, techniques, applications and sources},
  author={Wang, Xiao and Bo, Deyu and Shi, Chuan and Fan, Shaohua and Ye, Yanfang and Yu, Philip S},
  journal={IEEE transactions on big data},
  volume={9},
  number={2},
  pages={415--436},
  year={2022},
  publisher={IEEE}
}

@inproceedings{zhao2021heterogeneous,
  title={Heterogeneous graph structure learning for graph neural networks},
  author={Zhao, Jianan and Wang, Xiao and Shi, Chuan and Hu, Binbin and Song, Guojie and Ye, Yanfang},
  booktitle={AAAI},
  year={2021}
}

@inproceedings{kim2025hello,
  title={'Hello, World!': Making GNNs Talk with LLMs},
  author={Kim, Sunwoo and Lee, Soo Yong and Yoo, Jaemin and Shin, Kijung},
  booktitle={EMNLP Findings},
  year={2025}
}

\end{document}